\documentclass[10pt,twocolumn,letterpaper]{article}
\usepackage{multirow}
\usepackage{iccv}
\usepackage{times}
\usepackage{epsfig}
\usepackage{graphicx}
\usepackage{amsmath}
\usepackage{amssymb}

\usepackage{booktabs}
\usepackage{algorithm}
\usepackage{algorithmic}
\usepackage{times}
\usepackage{epstopdf}
\usepackage{epsfig}
\usepackage{graphicx}
\usepackage{amsmath}
\usepackage{amssymb}
\usepackage{booktabs}       
\usepackage{amsfonts}       
\usepackage{nicefrac}       
\usepackage{microtype}      
\usepackage{tabularx}
\usepackage{algorithm}
\usepackage{algorithmic}
\usepackage{epsfig}
\usepackage{amsmath}
\usepackage{multirow}
\usepackage{subfigure}
\usepackage{setspace}

\usepackage[breaklinks=true,bookmarks=false]{hyperref}

\iccvfinalcopy 

\def\iccvPaperID{****} 

\begin{document}

\title{GLMNet: Graph Learning-Matching Networks for Feature Matching}

\author{Bo Jiang, Pengfei Sun, Jin Tang and Bin Luo\\
Computer Science and Technology, Anhui University, China\\
{jiangbo@ahu.edu.cn}
}

\maketitle

\begin{abstract}
Recently, graph convolutional networks (GCNs) have shown great potential for the task of graph matching.
It can integrate graph node feature embedding, node-wise affinity learning and matching optimization
together in a unified end-to-end model.
One important aspect of graph matching is the construction of two matching graphs.
However, the matching graphs we feed to existing graph convolutional matching networks are generally fixed and independent of graph matching,
which thus are not guaranteed to be
optimal for the graph matching task. 
Also, existing GCN matching method employs several general smoothing-based graph convolutional layers to generate graph node embeddings, in which extensive smoothing convolution operation may dilute the desired discriminatory information of graph nodes.
To overcome these issues, we propose a novel Graph Learning-Matching Network (GLMNet) for graph matching problem.
GLMNet has three main aspects.
(1) It integrates graph learning into graph matching which thus adaptively learn a pair of optimal graphs that best serve graph matching task.
(2) It further employs a Laplacian sharpening convolutional module to generate more discriminative node embeddings for graph matching.
(3) A new constraint regularized loss is designed for GLMNet training which can encode the desired one-to-one matching constraints in matching optimization.
Experiments on two benchmarks demonstrate the effectiveness of GLMNet and advantages of its main modules.

\end{abstract}

\section{Introduction}

Many problems of interest in computer vision and pattern recognition area can be formulated as a problem of finding consistent
correspondences between two sets of features which is known as feature matching problem.
Feature set that incorporates the pairwise constraint can be represented via an attribute graph whose nodes represent the unary  descriptors of feature points and
edges encode the pairwise relationships among different feature points.
Based on this graph representation, feature matching can then be reformulated as graph node matching problem.

Graph matching generally first operates with both node and edge affinities that encode similarities between node and edge descriptors in two graphs. 
Then, it is can be formulated mathematically as an Integral Quadratic Programming (IQP) problem with permutation constraint on related solution
to encode the one-to-one matching constraints~\cite{yan2016short}. It is known to be NP-hard.
Thus, many methods normally solve it approximately by relaxing the discrete permutation constraint and finding local optimal solutions~\cite{Enqvist2009Optimal,yan2016short,FGM,RWGM,NOGM}.
In addition, to obtain better node/edge affinities, learning methods have been investigated to determine the more optimal parameters in node/edge
affinity computation~\cite{caetano2009learning,cho2013learning,Leordeanu2012Unsupervised}.
Recently, deep learning methods have also been developed for matching problem~\cite{nowak2018revised,zanfir2018deep,wang2019learning}.
The main benefit of deep learning matching methods is that
they can conduct visual feature representation, node/edge affinity learning and matching optimization together in an end-to-end manner.
Zanfir et al.,~\cite{zanfir2018deep} propose an end-to-end graph matching model which
makes it possible to learn all parameters of the graph matching process.
 Wang et al.,~\cite{wang2019learning} recently propose to explore graph convolutional networks (GCNs) for graph matching which
conducts graph node embedding and matching simultaneously in a unified network.

Inspired by recent deep graph matching methods, in this paper, we propose a novel Graph Learning-Matching Network (GLMNet)
for graph matching problem. Overall, the main contributions of this paper are three aspects. 

First, one important aspect of (feature) graph matching is the construction of two matching graphs.
Existing deep graph matching models~\cite{zanfir2018deep,wang2019learning} generally use fixed structure graphs, such as k-NN, Delaunary graph, etc.,
which thus are not guaranteed to best serve the matching task. To address this issue, we propose to
adaptively learn a pair of optimal graphs for the matching task and integrate \emph{graph learning} and
\emph{graph matching} simultaneously in a unified end-to-end network architecture.

Second, existing GCN based graph matching model~\cite{wang2019learning} adopts the general smoothing based graph convolution operation~\cite{kipf2016semi} for graph node embedding
which may encourage the feature embedding of each node becoming more similar to those of its neighboring nodes~\cite{li2018deeper}.
This is desirable for graph node labeling or classification tasks~\cite{kipf2016semi}, \emph{but} undesirable for the matching task because
 extensive smoothing convolution may dilute the discriminatory information.
To alleviate this affect, we propose to incorporate a Laplacian sharpening based graph convolution operation~\cite{park2019symmetric} for graph node embedding and matching task.
Laplacian sharpening process can be regarded as the counterpart of
Laplacian smoothing which encourages the embedding of each node farther away from its neighbors.

Third, existing deep graph matching methods generally utilize a doubly stochastic normalization for the final matching prediction~\cite{zanfir2018deep,wang2019learning}.
This generally ignores the discrete one-to-one matching constraints in matching optimization/prediction.
To overcome this issue, we develop a novel constraint regularized loss to further incorporate the one-to-one matching constraints in matching prediction.

Experimental results including ablation studies demonstrate the effectiveness of our GLMNet and  advantages of
devised components including graph learning-matching architecture, Laplacian sharpening convolution for discriminative embedding,
and constraint regularized loss to encode one-to-one matching constraints. 

\section{Related Works}

\subsection{Graph convolutional networks}

Recently, Graph Convolutional Networks (GCNs) have been widely studied  to deal with graph node embedding and learning~\cite{bruna2014spectral,kipf2016semi,velickovic2017graph,park2019symmetric}.
The main advantage of GCNs is that they provide an end-to-end learning which thus can be incorporated into some other specific deep learning architectures.
One can refer work~\cite{wu2019comprehensive} for more comprehensive review. Here, we briefly review some works that are related with our model in this paper.
By exploring the first-order approximation of spectral filters, Kipf et al.,~\cite{kipf2016semi} propose a simple Graph Convolutional Network (GCN) for
graph node representation and semi-supervised learning. 
Li et al.,~\cite{li2018deeper} interpret GCNs~\cite{kipf2016semi} from graph Laplacian smoothing and show that feature representations of graph nodes will become more similar as
network depth increases.
Recently, Park et al.,~\cite{park2019symmetric} introduce a novel Laplacian sharpening convolution operation and propose a symmetric graph convolutional
 autoencoder model for unsupervised graph representation learning.
To further incorporate graph learning into GCNs,
Veli{\v{c}}kovi{\'c} et al.,~\cite{velickovic2017graph} propose Graph Attention Networks (GATs) for graph based semi-supervised learning. 
 Li et al.,~\cite{velickovic2017graph} present an adaptive graph CNNs,
 in which the graph is learned adaptively via a metric learning method.
Jiang et al.,~\cite{jiang2019semi} propose Graph Learning-convolutional Network (GLCN) for graph node semi-supervised classification by
 integrating both graph learning and convolutional representation together in a unified network architecture.

\subsection{Deep graph matching}

Graph matching is a fundamental problem in computer vision and pattern recognition area and has been widely studied.
Recently, deep learning models have been developed for graph matching problem.
Nowak et al.,~\cite{nowak2018revised} propose to explore graph neural networks (GNNs) for solving the general Quadratic Assignment Problem (QAP)
which can be used for graph matching problem.
Li et al.,~\cite{li2019graph} propose Graph Matching Networks (GMNs) for learning the similarity of graph structured objects.
This work focuses on learning the similarity between two graphs. Differently, here we focus on graph node one-to-one matching problem. 
Zanfir et al.,~\cite{zanfir2018deep} propose an end-to-end graph matching model which integrates node feature extraction, node/edge affinities learning
and matching optimization together in a unified network.
Wang et al.,~\cite{wang2019learning} recently propose to explore graph convolutional networks (GCNs)  for graph matching task which
conducts graph node embedding and matching prediction simultaneously in a unified network.
The core of this method~\cite{wang2019learning} is to learn an optimal  embedding for graph matching task based on which the
final graph matching prediction can be approximately transferred as a linear assignment problem and thus can be solved via a Sinkhorn operation~\cite{adams2011ranking}. 


Following to this research direction, in this paper, we propose a new Graph Learning-Matching Network (GLMNet)
by further exploiting graph convolutional networks for graph matching task.
In contrast to previous works~\cite{zanfir2018deep,wang2019learning}, the main contributions of GLMNet are follows.
First, it incorporates \emph{graph learning} into graph matching network.
To our best knowledge, this is the first time to incorporate graph learning into graph matching to build an end-to-end learning network.
Second, GLMNet employs a more reasonable sharpening-based graph convolutional embedding for graph node embedding and matching task.
Third, in GLMNet, a new constraint regularized loss function is designed to encode the one-to-one matching constraints in graph matching optimization.

\begin{figure*}[ht]
\centering
\noindent\makebox[\textwidth][l] {
\includegraphics[width=1.0\textwidth]{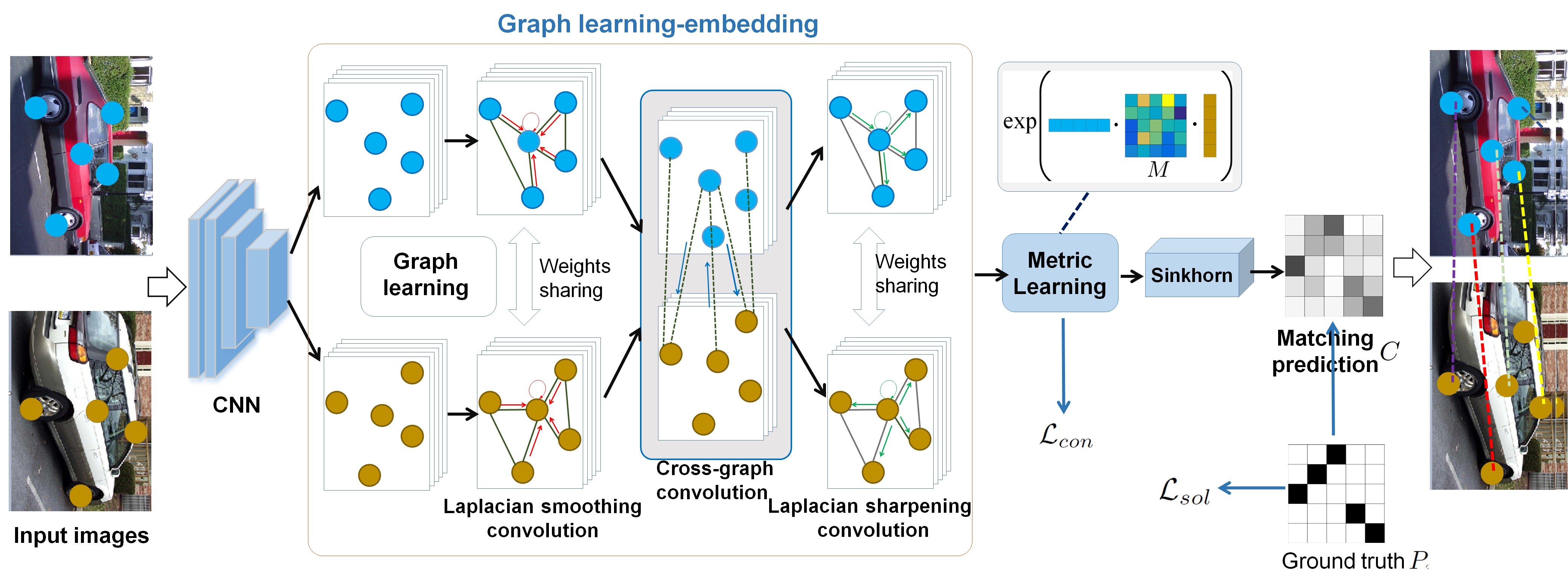}
}
\caption{Architecture of the proposed GLMNet which mainly contains node feature extraction, graph learning, graph convolutional embedding
and node affinity metric learning. The CNN model, graph learning and graph convolutional embedding and affinity metric are all
learnable in an end-to-end manner.}
\end{figure*}
\section{Proposed Approach}

\textbf{Problem Formulation}. Let $\mathcal{M}$ and $\mathcal{D}$ denote two feature sets of two images $\mathcal{I},\mathcal{I}'$, respectively.
The aim of feature matching is to determine the consistent correspondences between features of two images with one-to-one matching constraints (i.e., each feature in $\mathcal{M}$ can match at most one feature in $\mathcal{D}$ and vice versa).
To do so, for each feature in $\mathcal{M}$ and $\mathcal{D}$, we first extract a unary feature descriptor for it.
Let $X=(x_1, x_2\cdots x_m)$ and $Y=(y_1, y_2\cdots y_n)$ denote the collection of unary feature descriptors for these two sets,
 where $m, n$ denote the sizes of two feature sets, respectively.
Also, for each pair of feature points in $\mathcal{M}$ and $\mathcal{D}$, one can extract the binary relationship between them.
Therefore, one can build two graphs $G(X,A_x)$ and $G'(Y,G_y)$ for feature sets $\mathcal{M}$ and $\mathcal{D}$ whose nodes denote
the feature points attributed by feature descriptors $X, Y$ and edges $A_x, A_y$ encode the binary relationships between feature points.
Based on these representations, the task of feature matching can then be formulated as \emph{graph matching} that aims to determine the
 consistent correspondences between two graph node (feature) sets by considering both 1) how well the nodes' descriptors are matched
 and 2) how well the edges' attributes (relationships) are preserved~\cite{FGM,zanfir2018deep}.

One kind of popular approaches for graph matching problem is to utilize graph embedding based approaches that aim to first
 embed the nodes of two graphs into a common feature space and then utilize a metric learning technique to find the point correspondences in the feature space~\cite{caelli2004eigenspace,tang2012graph}.
 Comparing with original feature space $X, Y$, the node representations in the embedding space further
incorporate the information of graph structure and thus can become more discriminatively for the matching problem~\cite{tang2012graph,wang2019learning}.
 Wang et al.,~\cite{wang2019learning} recently propose to integrate graph embedding and matching optimization together via a supervised GCN architecture
 which can obviously boost their respectively performance.


\textbf{Overview of Approach}.
Inspired by recent work~\cite{wang2019learning}, our aim in this paper is to conduct the
 above node feature extraction, graph construction, graph node embedding and matching optimization together in
an end-to-end network framework.
We call our approach as Graph Learning-Matching Convolutional Network (GLMNet).
%
%
%
%
%
%
Figure 1 shows the overview of GLMNet which contains the following four modules. 
\begin{itemize}
  \item \emph{Deep feature extraction:} We utilize a CNN  to extract the  feature descriptors of all feature points for two matching images.
  \item \emph{Graph learning module:} We develop a graph learning module to adaptively learn a pair of optimal graphs  for graph matching problem.
  \item \emph{Graph convolutional embedding module:} We employ a novel GCN architecture to learn discriminative node embeddings
  for the node affinity learning and matching prediction.
  \item \emph{Affinity learning and matching prediction:} Based on the proposed graph convolutional embeddings, we finally conduct an affinity learning module  for matching prediction.
\end{itemize}

In the following sections, we present the details of these  modules respectively.

\subsection{Deeply learned node feature extraction}


In this paper, we focus on image feature (key-points, regions, etc) matching task. 
We adopt a CNN for their feature extraction, which are constructed by interpolating on CNN¡¯s feature map~\cite{wang2019learning}.
Specifically, we adopt a five-layer VGG-16 network~\cite{simonyan2014very} to extract a 1024 dimension feature descriptor for each keypoint/region. 
The parameters of the used VGG network are pre-trained on ImageNet~\cite{deng2009imagenet}.
In the following, we denote $X = (x_1,x_2 \cdots x_{m}) \in \mathbb{R}^{p\times m}$
and $Y = (y_1,y_2 \cdots y_{n})  \in \mathbb{R}^{p\times n}$ as the collections of feature descriptors for two feature sets, respectively. 

\subsection{Graph learning architecture}

One important aspect of graph matching is the feature-based graph construction $G(X, A_{x})$ and $G'(Y, A_{y})$. 
Constructing a good graph to represent feature relationships is generally important for graph convolutional embedding and matching tasks.
Traditional human established graphs  generally use fixed parameters to determine the graph structure and thus are not guaranteed to best serve the matching task.
To overcome this issue, we propose to learn a pair of optimal graphs $A_x, A_y$ adaptively and further provide a unified graph learning-matching network architecture
for the matching task. 

%

 Let $X=(x_1, x_2 \cdots x_m)\in \mathbb{R}^{n\times p}$ be $n$ data features.  
Our graph learning aims to seek a function $A_{x}(i,j) = \phi(x_i,x_j;\theta_x)$ with parameter $\theta_x$ to represent
 the pairwise relationship between data $x_i$ and $x_j$. Here, we implement $\phi(x_i,x_j;\theta_x)$ via a single-layer neural network,
 which is parameterized by a weight vector
$\theta_x$. Inspired by previous works~\cite{velickovic2017graph,jiang2019semi}, we propose to define $\phi(x_i,x_j;\theta)$ as
\begin{align}\label{}
 A_{x}(i,j)  = \phi(x_i,x_j;\theta_x) = \frac{\exp(\sigma( {\theta}_x^T[{x}_i||{x}_j]))}{\sum^n_{j=1}\exp(\sigma(\theta_x^T [{x}_i || {x}_j])}
\end{align}
where $||$ denotes the concatenation operation and $\sigma(\cdot)$ denotes an activation function, such as $\textrm{ReLU}(\cdot)=\max(0,\cdot)$. 
In some cases, when an initial graph ${A}'_x$ is available, we can thus incorporate it into the above graph learning as
\begin{align}\label{}
A_{x}(i,j)& = \phi(x_i,x_j,{A}'_x;\theta_x) \nonumber \\
&= \frac{\tilde{A}_{x}(i,j)\exp(\sigma( {\theta}_x^T[{x}_i||{x}_j]))}{\sum^n_{j=1}{A}'_{x}(i,j)\exp(\sigma(\theta_x^T [{x}_i||{x}_j]))}
\end{align}
In summary, we thus learn two optimal graphs for two feature sets respectively as,
\begin{align}\label{}
A_{x}(i,j)& = \phi(x_i,x_j,{A}'_x;\theta_x)  \\
A_{y}(i,j)& = \phi(y_i,y_j,{A}'_y;\theta_y)
\end{align}
where ${A}'_x$ and ${A}'_y$ denote the initial graphs for two feature sets respectively, and $\theta_x, \theta_y$ denote trainable parameters.
In our experiments, we set $\theta_x=\theta_y$ to encourage consistent graph learning across two feature sets~\cite{jiang2019unified}.

\subsection{Graph convolutional module}


%
As shown in Figure 1, our graph convolutional module involves two intra-graph convolutional submodules, i.e., smoothing convolution layer and
sharpening convolution layer, and one cross-graph convolutional submodule.

 \textbf{Intra-graph Convolutional module.}
The aim of intra-graph convolutional embedding is to
generate discriminative embeddings of graph nodes for matching problem by taking in considering both unary feature representations of nodes and binary relationships of edges.
Note that existing GCNs mostly adopt (Laplacian) smoothing based graph convolution operations in node representation which encourage the
latent representation of each node similar to those of its neighboring nodes as depth increases~\cite{li2018deeper}.
This is desirable for graph node labeling or classification tasks, but undesirable for the matching task.
Because 
extensive smoothing convolution may dilute the discriminatory information. To alleviate this affect,
inspired by recent work~\cite{park2019symmetric}, we propose to further incorporate Laplacian sharpening convolution and
employ both smoothing and sharpening convolution operations for graph node embeddings.


\noindent \textbf{\emph{Laplacian smoothing convolution}}.
Given input node embeddings  $\{{X}^{(k)}, {Y}^{(k)}\}$ of two graphs in the $k$-th hidden layer, we can obtain the optimal graphs $\{{A}_x^{(k)}, A^{(k)}_{y}\}$ by
using Eqs.(3,4). Then, Laplacian smoothing convolution aims to conduct the layer-wise propagation as
\begin{align}\label{EQ:layer_gcn0}
X^{(k+1)}& = \sigma\big[(1-\gamma)X^{(k)}\Theta_n^{(k)} + \gamma \tilde{A}^{(k)}_{x}X^{(k)}\Theta_e^{(k)}\big]  \\
Y^{(k+1)}& = \sigma\big[(1-\gamma)Y^{(k)}\Theta_n^{(k)} + \gamma \tilde{A}^{(k)}_{y}Y^{(k)}\Theta_e^{(k)}\big]
\end{align}
where $k=0,1\cdots K-1$ and $\sigma(\cdot)$ denotes an activation function, such as $\mathrm{ReLU}(\cdot) = \max(0,\cdot)$, and parameter $\gamma\in(0,1)$ balances two terms.
In our experiments, we set $\gamma=0.5$.
$\tilde{A}_x^{(k)}$ and $\tilde{A}_y^{(k)}$ denote the row-normalized Laplacian matrix\footnote{Given any matrix $A$, its row-normalized Laplacian is defined as $D^{-1}A$, where
$D$ is the diagonal matrix with $D_{ii}=\sum_j A_{ij}$} of ${A}_x^{(k)}, {A}_y^{(k)}$ respectively.
The parameters $\Theta^{(k)}=\{\Theta_n^{(k)},\Theta_e^{(k)}\}$ denote layer-specific trainable weight matrices.
Here, we use two trainable weight matrices $\{\Theta_n^{(k)},\Theta_e^{(k)}\}$ to learn node unary representation and
 propagation representation respectively, as suggested in work~\cite{wang2019learning}. 
The network parameters $\Theta^{(k)} = \{\Theta_n^{(k)}, \Theta_e^{(k)}\}$ are shared
across two graphs, which can encourage to learn consistent node embeddings across two graphs, as suggested in other works~\cite{wang2019learning,jiang2019unified}.

\noindent \textbf{\emph{Laplacian sharpening convolution}}.
To further enhance the discriminative ability of graph node embeddings, we also employ a Laplacian sharpening convolutional module in our GLMNet.
Laplacian sharpening can be regarded as the counterpart of Laplacian smoothing which encourages the embedding of each node farther away from its neighbors~\cite{park2019symmetric}.
Formally, given input node embeddings  $\{{X}^{(k)}, {Y}^{(k)}\}$ of two matching graphs in the $k$-th hidden layer, we can obtain the optimal graphs $\{{A}_x^{(k)}, A^{(k)}_{y}\}$ by
using Eqs.(3,4). Then, we propose to conduct Laplacian sharpening convolution as, 
\begin{align}\label{EQ:layer_gcn}
X^{(k+1)} & = \sigma\big[(1+\tilde{\gamma})X^{(k)}\Theta_n^{(k)} -\tilde{\gamma} \tilde{A}^{(k)}_{x}X^{(k)}\Theta_e^{(k)}\big]  \\
Y^{(k+1)}& = \sigma\big[(1+\tilde{\gamma})Y^{(k)}\Theta_n^{(k)} -\tilde{\gamma}\tilde{A}^{(k)}_{y}Y^{(k)}\Theta_e^{(k)}\big]
\end{align}
where parameter $\tilde{\gamma} > 0$ balances two terms and is set to 0.75 in our experiments.
$\tilde{A}_x^{(k)}$ and $\tilde{A}_y^{(k)}$ denote the row-normalized Laplacian matrix of ${A}_x^{(k)}, {A}_y^{(k)}$ respectively.
Similar to Eqs.(5,6), here we also use trainable parameter matrices $\Theta^{(k)}=\{\Theta_n^{(k)},\Theta_e^{(k)}\}$ to learn node unary representation and
 propagation representation respectively. These parameter matrices are shared across two graphs
to encourage to learn consistent node representations across two matching graphs.

 \textbf{Cross-graph Convolutional Module.}
Similar to work~\cite{wang2019learning}, we further leverage a cross graph convolutional learning module to mine
the correlations between the embeddings of two graphs. 
Given embeddings ${X}^{(k)}\in \mathbb{R}^{m\times d_k}$ and ${Y}^{(k)}\in \mathbb{R}^{n\times d_k}$ of two graphs, we first compute the co-affinity matrix $C^{(k)}_{xy}$ between them as,
\begin{align}\label{EQ:layer_cgcn}
C^{(k)}_{xy}(i,j) = \exp\Big(\frac{({{X}^{(k)}}^{T}W{Y}^{(k)})_{ij}}{\delta}\Big) \in \mathbb{R}^{m\times n}
\end{align}
where $W\in \mathbb{R}^{d_k\times d_k}$ is a trainable weight matrix.
We can also compute $C^{(k)}_{yx}$ similarly.
Based on these co-affinity matrices $\{C^{(k)}_{xy},C^{(k)}_{yx}\}$, we then conduct cross-graph convolutional learning as
\begin{align}\label{EQ:layer_gcn}
X^{(k+1)} = \big[ {C}^{(k)}_{xy}Y^{(k)} || X^{(k)} \big]\Theta_{xy}^{(k)} \\
Y^{(k+1)} = \big[ {C}^{(k)}_{yx}X^{(k)} || Y^{(k)} \big]\Theta_{yx}^{(k)}
\end{align}
where $||$ denotes the concatenation operation to incorporate the original feature information in cross graph convolution. Parameters $\Theta_{xy}=\{\Theta_{xy}^{(k)},\Theta_{yx}^{(k)}\}$
denote the layer-specific trainable weight matrices.

\subsection{Matching prediction and loss function}

\textbf{Affinity Metric Learning.}
Using the above graph convolutional embeddings, the final matching prediction
can be formulated as  node-to-node affinity metric learning in the embedding space.
Let $\widetilde{X}\in \mathbb{R}^{m\times d}, \widetilde{Y}\in \mathbb{R}^{n\times d}$ denote the output node embeddings of two graphs, respectively.
Then, the node-to-node affinity (similarity) matrix $\widetilde{C}$ can be learned as
\begin{align}\label{EQ:layer_cgcn}
\widetilde{C}(i,j) = \exp\Big(\frac{({\widetilde{X}}M\widetilde{Y}^{T})_{ij}}{\delta'}\Big)\in \mathbb{R}^{m\times n}
\end{align}
where $\widetilde{C}(i,j)$ denotes the similarity between node $i$
in the first graph $G$ and node $j$ in the second graph $G'$.
$M\in \mathbb{R}^{d\times d}$ denotes the learnable weight matrix of this affinity function.

For one-to-one matching problem, 
the ideal matching prediction $\widetilde{C}$ should satisfy the permutation constraint, i.e., there exists only one non-zero element in each row/column of matrix $\widetilde{C}$.
One possible way is to use a post-discretization operation (e.g.,Hungarian) on the learned $\widetilde{C}$.
However, the discretization operation is not differentiable, making the training of network more difficultly.
Thus, one can use the continuous Sinkhorn operation~\cite{sinkhorn1964relationship} to make the final matching prediction $\widetilde{C}$ satisfy the doubly-stochastic constraint, i.e.,
\begin{align}\label{EQ:layer_cgcn}
C = \mathrm{Sinkhorn}(\widetilde{C})
\end{align}
This Sinkhorn process has been shown effectively for permutation prediction and approximation~\cite{wang2019learning,adams2011ranking}.
Additionally, we will introduce a constraint regularized loss to further encourage the permutational matching prediction, as discussed below.

\textbf{Constraint Regularized Loss.}
 To further incorporate the one-to-one matching constraints, we develop a constraint regularized loss function to encourage the predicted matching solution satisfying the permutation constraint. 
 To do so, we first define an indicative matrix ${U}\in \mathbb{R}^{mn\times mn}$ which denotes the conflict relationships among different assignments/matches, i.e.,
\begin{equation}
{U}_{ij,kl} = \left\{ \begin{array}{ll}
1 & \textrm{if $i=k, j\neq l$ or $i\neq k, j=l$,}\\
0 & \textrm{otherwise.}
\end{array} \right.
\end{equation}
where $i,k=1, 2, \cdots m$ and $j,l = 1, 2, \cdots n$.
For the one-to-one matching problem, the ideal permutational matching solution $C$ should satisfy
\begin{align}
 \sum_{i,j}\sum_{k,l}{U}_{ij,kl}C_{ij}C_{kl} = 0
\end{align}
%
Note that, the above final output matching prediction $C$ is doubly stochastic and thus nonnegative.
The motivates us to develop the following constraint regularized loss to encourage the learned matching prediction satisfying the one-to-one matching constraints
 as much as possible,
\begin{equation}
\mathcal{L}_{con} = \sum_{i,j}\sum_{k,l}{U}_{ij,kl}C_{ij}C_{kl}
\end{equation}

 \textbf{Cross Entropy Loss.}
 For the matching prediction, we use cross entropy loss function.
Let $P\in \mathbb{R}^{m\times n}, P_{ij}\in \{0,1\}$ denotes the ground truth permutation matrix solution.
Then, we adopt the cross entropy loss to train our model.
The cross entropy loss is defined as~\cite{wang2019learning}
\begin{equation}
\mathcal{L}_{sol} = -\sum_{i,j} P_{ij}\log(C_{ij}) + (1-P_{ij})\log(1-C_{ij})
\end{equation}
Thus, the final overall loss function to train our GLMNet network in an end-to-end manner is formulated as
\begin{equation}
\mathcal{L} = \mathcal{L}_{sol} + \lambda \mathcal{L}_{con}
\end{equation}
where $\lambda >0 $ balances two terms and is set to 0.1 in our experiments.

%
\begin{figure*}[!htpb]
\centering
\noindent\makebox[\textwidth][l] {
\includegraphics[width=1.0\textwidth]{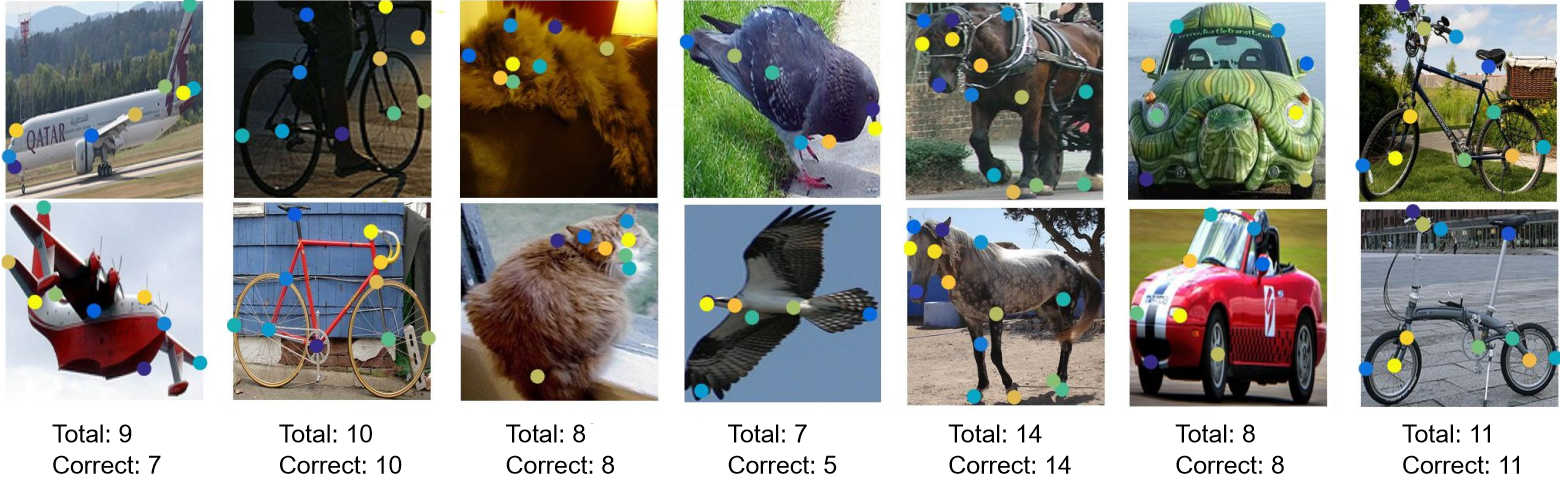}
}
\caption{Some matching examples of  GLMNet on PASCAL VOC test-set.  Colors identify the predicted matching
between key-points. Note that, GLMNet can obtain the correct matches for image pairs that are with large appearance and pose changes.}
\end{figure*}
%
\section{Experiments}


In this section, we evaluate the effectiveness of the proposed GLMnet on two benchmark datasets (PASCAL VOC~\cite{everingham2010pascal,bourdev2009poselets}, WILLOW-ObjectClass~\cite{cho2013learning}) and compare it with other competing methods including HARG-SSVM~\cite{cho2013learning}, GMN~\cite{zanfir2018deep}, GMN-PL~\cite{zanfir2018deep,wang2019learning}, PIA-GM-OL~\cite{wang2019learning}, PIA-GM~\cite{wang2019learning} and PCA-GM~\cite{wang2019learning}. 

\subsection{Network architecture and parameter setting}
\begin{table*}[t]
\caption{Comparison results of different methods on Pascal VOC dataset. Results indicated with * are taken from ~\cite{wang2019learning}. The best results are marked as bold. } \label{b1}
\begin{center}
\resizebox{\textwidth}{!}{
\LARGE
\begin{tabular}{l|c|c|c|c|c|c|c|c|c|c|c|c|c|c|c|c|c|c|c|c|c}
\hline
\hline
Method & areo & bike & bird & boat & bottle & bus & car &cat& chair&  cow & table & dog & horse & mbike & person & plant & sheep & sofa & train & tv & \textbf{mean} \\
\hline
GMN*&31.9 &47.2 &51.9 &40.8& 68.7& 72.2& 53.6 &52.8& 34.6& 48.6& 72.3& 47.7 &54.8 &51.0 &38.6 &75.1& 49.5& 45.0& 83.0 &86.3  & 55.3\\
GMN-PL*&31.1& 46.2& 58.2& 45.9& 70.6& 76.4& 61.2& 61.7& 35.5 &53.7& 58.9& 57.5& 56.9& 49.3& 34.1& 77.5& 57.1& 53.6& 83.2& 88.6&57.9\\
PIA-GM-OL*&39.7& 57.7& 58.6& 47.2& 74.0& 74.5& 62.1& 66.6& 33.6& 61.7& 65.4& 58.0& 67.1& 58.9& 41.9& 77.7& 64.7& 50.5& 81.8& 89.9&61.6\\
PIA-GM*&41.5&55.8& 60.9& 51.9& 75.0& 75.8& 59.6& 65.2& 33.3& 65.9& 62.8& 62.7& 67.7& 62.1& 42.9& 80.2& 64.3& 59.5& 82.7& 90.1&63.0\\
PCA-GM* &40.9& 55.0 &\textbf{65.8}& 47.9 &76.9 &\textbf{77.9}& 63.5 &67.4 &33.7 &65.5 &63.6 &61.3& \textbf{68.9}& 62.8& \textbf{44.9}& 77.5 &67.4 &57.5 &\textbf{86.7}& 90.9 & 63.8\\
\hline
GLMNet & \textbf{52.0}& \textbf{67.3} & 63.2 & \textbf{57.4} &\textbf{80.3}& 74.6 &\textbf{70.0} &\textbf{72.6} &\textbf{38.9}& \textbf{66.3} &\textbf{77.3} &\textbf{65.7} &67.9 &\textbf{64.2} &44.8 &\textbf{86.3} &\textbf{69.0}& \textbf{61.9 }&79.3&\textbf{ 91.3}&\textbf{67.5}\\
\hline
\hline
\end{tabular}}
\end{center}
\end{table*}

For feature extraction module, we adopt a deep convolutional neural network (VGG-16)~\cite{simonyan2014very} to extract image features, which is pre-trained on ImageNet~\cite{deng2009imagenet}.
More specifically, we extract features from relu4\underline{\hspace{0.5em}}2 and relu5\underline{\hspace{0.5em}}1 for fair comparison with
previous works~\cite{wang2019learning,zanfir2018deep}.
We concatenate these two kind of features together to form a 1024 dimension feature representation.
Our embedding network consists of three convolutional layers including one graph learning-smoothing convolutional layer, one cross-graph convolutional layer and one graph learning-sharpening convolutional layer.
The final embedding vectors of all nodes in two graphs are 2048 dimension.
The balancing parameters $\gamma, \tilde{\gamma}$ in GLMNet are fixed and set to $0.5,0.75$ respectively on all datasets.


\subsection{Evaluation on PASCAL VOC dataset}
%

Our first evaluation is performed on PASCAL VOC dataset~\cite{everingham2010pascal}  with Berkeley annotations of keypoints~\cite{bourdev2009poselets}.
This annotated dataset consists of 20 different categories of images which are varying
from its scale, pose and illumination.
Each image is annotated with 6$\sim$23 inlier keypoints.
Following the setting of previous works~\cite{wang2019learning},
we use 7020 annotated images which involve all 20 categories for  training and 1682 images for testing.
All images are cropped around its bounding box and resized to 256 $\times$ 256 before feeding to
our network for training and testing.
Table 1 summarizes the comparison results on this dataset.
The results of  other comparison methods have been reported in work~\cite{wang2019learning} and here we use them directly.
From Table 1, one can note that, the proposed GLMNet obviously outperforms other recent methods~\cite{wang2019learning,zanfir2018deep}.
Overall, GLMNet gains 3.7$\%$ improvement over PCA-GM~\cite{wang2019learning} which is also adopting GCN for graph matching and thus most related  with our GLMNet.
 This clearly demonstrates the effectiveness and advantage of the proposed GLMNet on solving graph based feature matching problem.
Also, GLMNet generally performs better than the other comparison methods on most image categories, indicating the robustness of GLMNet.
Figure 2 shows some matching examples on this dataset. One can note that, GLMNet can obtain the correct matches for image pairs that are with large appearance and pose changes.

\subsection{Evaluation on WILLOW-ObjectClass dataset}

This dataset contains object class images which are selected from Caltech-256 (Face, Duck, and Wine bottle) and PASCAL VOC2007 (Motorbike and Car)~\cite{everingham2010pascal} datasets. The images of these classes are selected such that each class contains at least 40 images.
Similar to PASCAL VOC dataset~\cite{everingham2010pascal}, each image is annotated with 10 keypoints. We also resize each image into 256 $\times$ 256 before feeding to our matching network.
Following the experiment setting~\cite{wang2019learning}, we first filter out the overlapping images in PascalVOC. Then, we initialize model weights by training the network on Pascal VOC Keypoint dataset~\cite{bourdev2009poselets}. They are later fine-tuned on the Willow dataset which denote as GMLNet-Willow. 
Table 2 summarizes the comparison results on this dataset.
For comparison methods HARG-SSVM~\cite{cho2013learning}, GMN-Willow~\cite{zanfir2018deep} and PCA-GM-Willow~\cite{wang2019learning},
we list the results on this dataset that have been reported in previous work~\cite{wang2019learning}. 
From Table 2, one can note that, the proposed GLMNet performs better than the other comparison related methods on this dataset. 
Overall, GMLNet-Willow gains 2.4$\%$ and 11.5$\%$ improvements over PCA-GM-Willow~\cite{wang2019learning} and GMN-Willow~\cite{zanfir2018deep}, respectively. 
 This further demonstrates the effectiveness of the proposed GLMNet for graph matching problem. 


\begin{table}[h]
\caption{Comparison results of different methods on WILLOW-ObjectClass dataset. Results indicated with * are taken from ~\cite{wang2019learning}. The best results are marked as bold. } \label{b1}
\begin{center}
\resizebox{238pt}{29pt}{
\begin{tabular}{ c|c|c|c|c|c|c }
\hline
\hline
Method & face& mbike& car& duck& bottle & \textbf{mean} \\
\hline
HARG-SSVM*~\cite{cho2013learning} & 91.2& 44.4& 58.4& 55.2& 66.6 &63.2\\
GMN-Willow*~\cite{zanfir2018deep} & 99.3 &71.4 &74.3& 82.8 &76.7 &80.9\\
PCA-GM-Willow*~\cite{wang2019learning} & 100.0& 76.7& 84.0& \textbf{93.5}& \textbf{96.9}&90.2\\
GLMNet-Willow & \textbf{100.0}& \textbf{89.7}& \textbf{93.6}& 85.4& 93.4& \textbf{92.4}\\
\hline
\hline
\end{tabular}}
\end{center}
\end{table}

\subsection{Ablation studies}

To justify the effectiveness of three main components (graph learning, Laplacian sharpening convolution and constraint regularization loss $\mathcal{L}_{con}$) in the proposed GLMNet model, we conduct ablation experiments on PASCAL VOC dataset~\cite{everingham2010pascal}.
Table 3 summarizes the ablation study results, where
tick in the table denotes the module is used in our model.
For comparison, we also report the result of PCA-GM~\cite{wang2019learning} as the baseline.
Here, we can note that (1) graph learning
module can significantly improve the matching results which clearly indicates the
advantage of the proposed graph learning architecture to adaptively learn optimal graphs for graph matching problem. 
(2) Incorporating Laplacian sharpening convolution is
 beneficial for learning more discriminative node embeddings which thus obviously improves the final
 matching accuracy. 
(3) The proposed constraint regularization loss is useful to
guide the more accurate graph matching prediction by further incorporating the desired one-to-one matching constraints in matching optimization.

\begin{table}[h]
\caption{Result of ablation studies. } \label{b1}
\begin{center}
\resizebox{238pt}{38pt}{
\begin{tabular}{c|c|c|c|c}
\hline
\hline
\multirow{2}{*}{PCA-GM} & \multirow{2}{*}{Graph learning} & \multirow{2}{*}{\begin{tabular}[c]{@{}l@{}}Constraint\\  loss $\mathcal{L}_{con}$\end{tabular}} & \multirow{2}{*}{\begin{tabular}[c]{@{}l@{}}Laplacian \\sharpening\end{tabular}} & \multirow{2}{*}{Accuracy} \\
                        &                            &                                                                              &                                                                                  &                           \\ \hline
 &\checkmark &\checkmark &\checkmark &\textbf{67.5}\\
 &\checkmark &\checkmark &$\times$ &66.9\\
 &\checkmark &$\times$ &$\times$ &66.6\\
\checkmark &$\times$ &$\times$ &$\times$ &63.8\\
\hline
\hline
\end{tabular}}
\end{center}
\end{table}

\section{Conclusion and Future Works}

This paper proposes a novel Graph Learning-Matching Network (GLMNet) model for graph matching.
GLMNet integrates  graph learning  and graph matching architectures together in a unified end-to-end network,
which can learn a pair of  optimal matching graphs that best serve the task of graph matching.
GLMNet employs a Laplacian sharpening convolution to  generate more discriminative node embeddings for matching task.
The proposed constraint regularized loss is further designed for GLMNet training to encode one-to-one matching constraints.
Experimental results on two benchmarks demonstrate that GLMNet obviously outperforms recent fixed-graph deep graph matching methods. 

Note that, GLMNet is not limited to deal with the task of two graph matching. In the future, we will adapt
GLMNet to address the more general multiple graph matching task.

{\small
\bibliographystyle{ieee}
\bibliography{egbib1}

\begin{thebibliography}{10}\itemsep=-1pt

\bibitem{adams2011ranking}
R.~P. Adams and R.~S. Zemel.
\newblock Ranking via sinkhorn propagation.
\newblock {\em arXiv preprint arXiv:1106.1925}, 2011.

\bibitem{bourdev2009poselets}
L.~Bourdev and J.~Malik.
\newblock Poselets: Body part detectors trained using 3d human pose
  annotations.
\newblock In {\em 2009 IEEE 12th International Conference on Computer Vision},
  pages 1365--1372. IEEE, 2009.

\bibitem{bruna2014spectral}
J.~Bruna, W.~Zaremba, A.~Szlam, and Y.~LeCun.
\newblock Spectral networks and locally connected networks on graphs.
\newblock In {\em International Conference on Learning Representations}, 2014.

\bibitem{caelli2004eigenspace}
T.~Caelli and S.~Kosinov.
\newblock An eigenspace projection clustering method for inexact graph
  matching.
\newblock {\em IEEE transactions on pattern analysis and machine intelligence},
  26(4):515--519, 2004.

\bibitem{caetano2009learning}
T.~S. Caetano, J.~J. McAuley, L.~Cheng, Q.~V. Le, and A.~J. Smola.
\newblock Learning graph matching.
\newblock {\em IEEE transactions on pattern analysis and machine intelligence},
  31(6):1048--1058, 2009.

\bibitem{cho2013learning}
M.~Cho, K.~Alahari, and J.~Ponce.
\newblock Learning graphs to match.
\newblock In {\em Proceedings of the IEEE International Conference on Computer
  Vision}, pages 25--32, 2013.

\bibitem{RWGM}
M.~Cho, J.~Lee, and K.~M. Lee.
\newblock Reweighted random walks for graph matching.
\newblock In {\em European Conference on Computer Vision}, pages 492--505,
  2010.

\bibitem{deng2009imagenet}
J.~Deng, W.~Dong, R.~Socher, L.-J. Li, K.~Li, and L.~Fei-Fei.
\newblock Imagenet: A large-scale hierarchical image database.
\newblock In {\em 2009 IEEE conference on computer vision and pattern
  recognition}, pages 248--255, 2009.

\bibitem{Enqvist2009Optimal}
O.~Enqvist, K.~Josephson, and F.~Kahl.
\newblock Optimal correspondences from pairwise constraints.
\newblock In {\em IEEE International Conference on Computer Vision}, pages
  1295--1302, 2009.

\bibitem{everingham2010pascal}
M.~Everingham, L.~Van~Gool, C.~K. Williams, J.~Winn, and A.~Zisserman.
\newblock The pascal visual object classes (voc) challenge.
\newblock {\em International journal of computer vision}, 88(2):303--338, 2010.

\bibitem{jiang2019unified}
B.~Jiang, X.~Jiang, A.~Zhou, J.~Tang, and B.~Luo.
\newblock A unified multiple graph learning and convolutional network model for
  co-saliency estimation.
\newblock In {\em Proceedings of the 27th ACM International Conference on
  Multimedia}, pages 1375--1382, 2019.

\bibitem{NOGM}
B.~Jiang, J.~Tang, C.~H. Ding, and B.~Luo.
\newblock Nonnegative orthogonal graph matching.
\newblock In {\em AAAI}, pages 4089--4095, 2017.

\bibitem{jiang2019semi}
B.~Jiang, Z.~Zhang, D.~Lin, J.~Tang, and B.~Luo.
\newblock Semi-supervised learning with graph learning-convolutional networks.
\newblock In {\em Proceedings of the IEEE Conference on Computer Vision and
  Pattern Recognition}, pages 11313--11320, 2019.

\bibitem{kipf2016semi}
T.~N. Kipf and M.~Welling.
\newblock Semi-supervised classification with graph convolutional networks.
\newblock {\em arXiv preprint arXiv:1609.02907}, 2016.

\bibitem{Leordeanu2012Unsupervised}
M.~Leordeanu and M.~Hebert.
\newblock Unsupervised learning for graph matching.
\newblock {\em International Journal of Computer Vision}, 96(1):28--45, 2012.

\bibitem{li2018deeper}
Q.~Li, Z.~Han, and X.-M. Wu.
\newblock Deeper insights into graph convolutional networks for semi-supervised
  learning.
\newblock In {\em Thirty-Second AAAI Conference on Artificial Intelligence},
  2018.

\bibitem{li2019graph}
Y.~Li, C.~Gu, T.~Dullien, O.~Vinyals, and P.~Kohli.
\newblock Graph matching networks for learning the similarity of graph
  structured objects.
\newblock In {\em International Conference on Machine Learning}, pages
  3835--3845, 2019.

\bibitem{nowak2018revised}
A.~Nowak, S.~Villar, A.~S. Bandeira, and J.~Bruna.
\newblock Revised note on learning quadratic assignment with graph neural
  networks.
\newblock In {\em 2018 IEEE Data Science Workshop (DSW)}, pages 1--5, 2018.

\bibitem{park2019symmetric}
J.~Park, M.~Lee, H.~J. Chang, K.~Lee, and J.~Y. Choi.
\newblock Symmetric graph convolutional autoencoder for unsupervised graph
  representation learning.
\newblock In {\em Proceedings of the IEEE International Conference on Computer
  Vision}, pages 6519--6528, 2019.

\bibitem{simonyan2014very}
K.~Simonyan and A.~Zisserman.
\newblock Very deep convolutional networks for large-scale image recognition.
\newblock {\em arXiv preprint arXiv:1409.1556}, 2014.

\bibitem{sinkhorn1964relationship}
R.~Sinkhorn.
\newblock A relationship between arbitrary positive matrices and doubly
  stochastic matrices.
\newblock {\em The annals of mathematical statistics}, 35(2):876--879, 1964.

\bibitem{tang2012graph}
J.~Tang, B.~Jiang, A.~Zheng, and B.~Luo.
\newblock Graph matching based on spectral embedding with missing value.
\newblock {\em Pattern Recognition}, 45(10):3768--3779, 2012.

\bibitem{velickovic2017graph}
P.~Velickovic, G.~Cucurull, A.~Casanova, A.~Romero, P.~Lio, and Y.~Bengio.
\newblock Graph attention networks.
\newblock {\em arXiv preprint arXiv:1710.10903}, 2017.

\bibitem{wang2019learning}
R.~Wang, J.~Yan, and X.~Yang.
\newblock Learning combinatorial embedding networks for deep graph matching.
\newblock In {\em ICCV}, 2019.

\bibitem{wu2019comprehensive}
Z.~Wu, S.~Pan, F.~Chen, G.~Long, C.~Zhang, and P.~S. Yu.
\newblock A comprehensive survey on graph neural networks.
\newblock {\em arXiv preprint arXiv:1901.00596}, 2019.

\bibitem{yan2016short}
J.~Yan, X.-C. Yin, W.~Lin, C.~Deng, H.~Zha, and X.~Yang.
\newblock A short survey of recent advances in graph matching.
\newblock In {\em Proceedings of the 2016 ACM on International Conference on
  Multimedia Retrieval}, pages 167--174, 2016.

\bibitem{zanfir2018deep}
A.~Zanfir and C.~Sminchisescu.
\newblock Deep learning of graph matching.
\newblock In {\em Proceedings of the IEEE Conference on Computer Vision and
  Pattern Recognition}, pages 2684--2693, 2018.

\bibitem{FGM}
F.~Zhou and F.~D. la~Torre.
\newblock Factorized graph matching.
\newblock In {\em IEEE Conference on Computer Vision and Pattern Recognition},
  pages 127--134, 2012.

\end{thebibliography}
}

\end{document}